# Guided parallelized stochastic gradient descent for delay compensation

Anuraganand Sharma

School of Computing, Information and Mathematical Sciences

The University of the South Pacific, Fiji

sharma_au@usp.ac.fj

## Abstract

Stochastic gradient descent (SGD) algorithm and its variations have been effectively used to optimize neural network models. However, with the rapid growth of big data and deep learning, SGD is no longer the most suitable choice due to its natural behavior of sequential optimization of the error function. This has led to the development of parallel SGD algorithms, such as asynchronous SGD (ASGD) and synchronous SGD (SSGD) to train deep neural networks. However, it introduces a high variance due to the delay in parameter (weight) update. We address this delay in our proposed algorithm and try to minimize its impact. We employed guided SGD (gSGD) that encourages consistent examples to steer the convergence by compensating the unpredictable deviation caused by the delay. Its convergence rate is also similar to A/SSGD, however, some additional (parallel) processing is required to compensate for the delay. The experimental results demonstrate that our proposed approach has been able to mitigate the impact of delay for the quality of classification accuracy. The guided approach with SSGD clearly outperforms sequential SGD and even achieves the accuracy close to sequential SGD for some benchmark datasets.

Keywords: asynchronous/synchronous stochastic gradient descent, classification, deep learning, Gradient Methods, Stochastic gradient descent.

## 1. Introduction

Knowledge discovery from data with machine learning techniques has been widely in use for several applications in medicine, economics, engineering, etc. (Kourou et al., 2015; Qiu et al., 2016; Schmidt et al., 2019). However, the rapid expansion of data (or big data) also poses a challenge for machine learning techniques to handle a vast amount of data in a limited time. Traditionally these learning techniques were confined to centralized processing, but recently, there has been enough attention given in to equip traditional techniques to speed up the training process in parallel on multiple machines/CPUs/GPUs. There are many machine learning approaches but we focus on gradient descent (GD) optimization based learning systems in this paper.

There are some sophisticated parallel machine learning processing techniques for big data such as MapReduce ("MapReduce - an overview | ScienceDirect Topics," 2019) and GraphLab (Low et al., 2010) that use exact algorithm rather than any approximation. Here the computation of partial gradients for weights is done in parallel to have faster conversion (Chu et al., 2007). The parallelization of gradient computation for GD, as discussed in this paper, may further speed-up the processing time. The other popular choice is ensemble learning where independent models are trained in parallel, however, they improve the accuracy but not the speed (Brown et al., 2005; Dean et al., 2012). Ensemble learning with Neural Networks (NN) is to train multiple models on the same dataset instead of a single model and to combine the predictions from these models to make a final outcome or prediction. This approach not only reduces the variance of neural network models but also results in predictions that are better than any single model. Training NN relies on optimization algorithms for convergence and GD and its variations are considered de-facto standard for parameter optimization for NN. To increase the processing speed in NN architecture the current trend is to parallelize the GD algorithms (Dean et al., 2012; Zheng et al., 2017). However, these GD techniques are inherently sequential (recursive on dependent parameters) that brings some degree of uncertainty with these approaches.

There has also been plethora of gradient based second-order distributed optimization algorithm such as DINGO (Crane and Roosta, 2019), DiSCO (Zhang and Xiao, 2015) and GIANT (Wang et al., 2018). However, they generally have either higher computational cost or have reliance on function convexity and choice of parameters (Crane and Roosta, 2019). The focus of this paper is only Sequential Gradient Descent (SGD) – a first-order GD algorithm – and its variants. SGD is considered a default standard optimization algorithm for many gradient based machine learning classification models such as neural networks and logistic regression because of its simplicity, fast convergence and applicability for non-convex functions (Lei et al., 2019).

Recently some attempts have been made to parallelize the gradient computation to achieve faster training speed. Algorithms such as parallelized SGD (Zinkevich et al., 2010), distributed stochastic optimization (Agarwal and Duchi, 2011), downpour SGD – an Asynchronous SGD (ASGD) technique (Dean et al., 2012) and Delay Compensation ASGD (DC-ASGD) (Zheng et al., 2017) have been developed. Notably, the gradient computation and parameter update in parallel do not make much of a mathematical sense (Zheng et al., 2017). Since by the time gradient is updated in anticipation to move towards the descent, its direction may no longer be pointing towards the expected descent. This scenario is explained in Fig. 1 where Part (a) shows that Parallel SGD finds the gradient of four mini-batches at time $t$ shown as arrows on dotted curves at $W_t$ with average error as the smooth curve. Part (b) shows the impact of delay or "long jump" when $W_t$ is updated. Mini-batches $1-3$ are favorable for error reduction even with the delay as the direction of the gradients correspond to the true gradient. However, mini-batch 4 suffers from delay in update at $W_{t+4}$ as its gradient still points to the direction what it was pointing to at $W_t$. Gradient only tells the direction of descent in nearby region, not far away regions. It may be referring to a totally different trough. We have termed this delay as "long jump" which at times may produce unrealistic solutions. The major issue would be when a jump is too long which would result in unpredictable solutions. So we must use "small" jump, however, this would defeat the purpose of parallel computation to make full use of resources. So we must set the threshold of an appropriate value to produce optimum results. Some research has been done to compensate for this "delay" so that maximum efficiency is attained with a small loss of accuracy. For example, it was reported in (Zheng et al., 2017) that DC-ASGD maintains the same efficiency as ASGD and achieves much higher accuracy due to its delay compensation approach. Stich et al. in (Stich et al., 2018) proposed an approach for setting up of distributed training with SGD to the other workers that combines gradient quantization with an error compensation technique. This scheme of sparsification/compression reduces the bottleneck of communication overhead which also converges at the same rate as vanilla SGD.

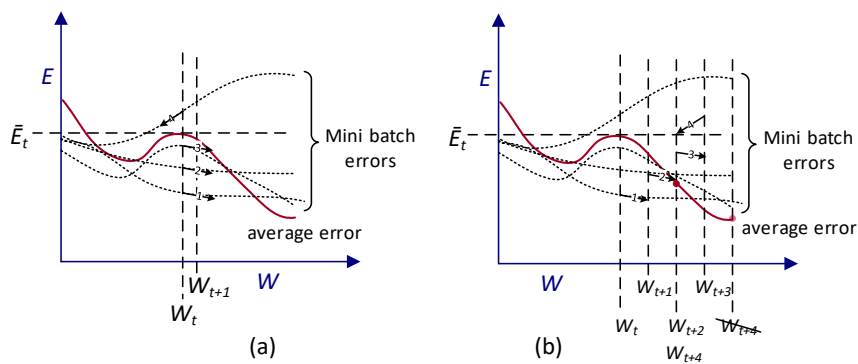

**Fig. 1.** "long jump" issues – delayed update

The remainder of this paper is organized as follows: Section 2 describes the structure of the parallel SGD algorithm. Section 3 has been allocated for convergence analysis of synchronous and asynchronous SGD algorithms in parallel processing framework. Section 4 describes the main contribution of this paper – the guided approach for delay compensation. Section 5 discusses the experiments and results on parallel and sequential SGD algorithm on nine benchmark datasets. Lastly, Section 6 concludes the paper with some suggestions for future work.

## 2. Parallel Gradient Descent

The natural behavior of the GD algorithm is sequential where the weight parameter $W_t$ is iteratively updated such that at each step a short distance is moved in the direction of error's rate of descent for each data instance (Bishop, 1995; Howard Anton et al., 2012). A simple mini-batch gradient descent algorithm is shown in Fig. 2. GD algorithm has only two major steps: gradient computation and weight update. The algorithm begins with the initialization of learning rate $\eta$, initial weight vector $W_0$ and total iterations $T$. $\eta$ is generally fixed between 0 – 1 but adaptive learning rates are also being used (Moreira and Fiesler, 1995; Yu and Liu, 2002). The size of the training dataset is $N$ with a data sample $d_i$. $\nabla E(W_t, d_i)$ is the gradient of error function $E(W_t, d_i)$ and $m$ is the size of the mini-batch. The algorithm stops when the termination criteria is met which is generally the given maximum iterations, or just before the overfitting starts (Michalewicz and Fogel, 2004).

```
Initialize η and W₀
For t = 1 … T
    Gradient computation
    ∇E̅ = 0;
    For k ∈ {1 … m}
        ∇E̅ = ∇E̅ + ∇E(W_{t-1}, d_k)
    End for
    ∇E̅ = ∇E̅/m
    Update weight vector
    W_t = W_{t-1} − η∇E̅
    t = t + 1
    Check Termination criteria
End for
```

**Fig. 2.** A simple min-batch gradient descent

Since a GD algorithm is not parallel by nature due to its dependency on sequential weight update, some approaches have been suggested by (Dean et al., 2012; Zheng et al., 2017; Stich et al., 2018) that compensates the parallelization of the weight update. A general approach is to either synchronously or asynchronously do the weight update in parallel that will have its own pros and cons. Asynchronous means no lock so any node can update the weight vector. That means other machines either use the new weight or stale weight depends on when they are fetched. The synchronous (i.e. with locks) SGD (SSGD) would ensure all nodes would get fresh weight, however, it suffers from latency by the slowest node. All other nodes must wait for the slowest node to complete the process because the parameter server is locked until all grade computations are done (Dean et al., 2012; Sridhar, 2015). Therefore, the synchronous approach is slower than the asynchronous approach but produces marginally better results as shown in (Zheng et al., 2017) and the experimental results and convergence analysis in this paper.

The algorithm for SSGD is shown in Fig. 3 which is basically the SGD algorithm where gradient computation happens in parallel. The gradient is then transferred to the parameter server synchronously where the weight update happens. Since the parameter server waits for the gradients to arrive, it results in a delay in parameter update. If the locks are removed it becomes asynchronous SGD (ASGD) (Zheng et al., 2017). The algorithm is shown in Fig. 4.

```
FOR t = 1:T
    W_t = READ W_t from the parameter server;
    d = GET data sample from data Storage;
    v_t(d) = calcGradient(W_t, d);
    LOCK parameter server
        SEND gradient to the parameter server;
    RELEASE LOCK
END FOR
```

**Fig. 3.** Parallel synchronous gradient descent algorithm

```
Listen
    If a gradient received
        W_t = W_t − ηv_t(d)
    End If
While (t = 1: T)
```

**Fig. 4.** Algorithm for the parameter server

## 3. Convergence Analysis

Basically, the GD algorithm is an optimization problem that can be represented as:

$$\widehat{W} = \underset{W}{\text{argmin}} \frac{1}{N} \sum_{n=1}^{N} E(d_n, W) \qquad (1)$$

where $E(d_n, W)$ is an error or cost function that takes weight vector $W$ and $N$ mini-batches of training examples of size $m$ from $d_1 \ldots d_N$. In a parallel architecture, there are multiple CPUs/cores where $c$ gradients are calculated in parallel in $c$ slave threads with different data mini-batches $d_i$ which are then queued to the parameter server to update $W$. The gradient computation can be shown as $\eta v_t(d_{i+1}), \eta v_t(d_{i+2}), \ldots, \eta v_t(d_{i+c})$ where $v_t(d_i)$ is a partial gradient function for a data mini-batch $d_i$ at iteration $t$.

If the gradient threads wait for other threads to complete their work, then it will be synchronous GD as the next iteration will depend on the slowest thread to complete the gradient computation. In this case, the weight update at iteration $(t+1)$ would be:

$$W_{t+1} = W_t - \eta v_t(d_{i+1}) - \eta v_t(d_{i+2}) - \cdots - \eta v_t(d_{i+c}) = W_t - \eta \sum_{j=1}^{c} v_t(d_{i+j})$$

Or, $W_{t+1} = W_t - \eta c \bar{v}_t$ where expected value $\mathbb{E}(\bar{v}_t) = \nabla E(W_t)$. $\nabla E(W_t)$ shows the true gradient for $W_t$.

In a simplistic case, the error function $E$ is convex and Lipschitz β-smooth (Lacoste-Julien et al., 2012; Bubeck, 2015) as shown in Eq. (2), i.e.,

$$E(W_{t+1}) \leq E(W_t) + \langle \nabla E(W_t), W_{t+1} - W_t \rangle + \frac{\beta}{2} \| W_{t+1} - W_t \|^2 \qquad (2)$$

Or, $E(W_{t+1}) \leq E(W_t) + \langle \nabla E(W_t), -\eta c \bar{v}_t \rangle + \frac{\beta}{2} \| -\eta c \bar{v}_t \|^2$

$$= E(W_+) + \langle \nabla E(W_t), W_t - W_+ \rangle - \eta c \langle \nabla E(W_t), \bar{v}_t \rangle + \frac{\beta \eta^2 c^2}{2} \| \bar{v}_t \|^2$$

where $W_+ = \underset{W}{\text{argmin}}\, E(W)$. This can be resolved to:

$\mathbb{E}(E(W_{t+1})) \leq \mathbb{E}(E(W_+) + \langle \nabla E(W_t), W_t - W_+ \rangle) - \eta c \| \nabla E(W_t) \|^2 + \frac{\eta c}{2} ( \| \nabla E(W_t) \|^2 + Var(\bar{v}_t))$

$$= \mathbb{E}(E(W_+) + \langle \nabla E(W_t), W_t - W_+ \rangle) - \frac{\eta c}{2} \|\nabla E(W_t)\|^2 + \frac{\eta}{2}\sigma^2$$

where $Var$ refers to variance. Here we assume $\beta \leq \frac{1}{\eta c}$ and $Var(\bar{v}_t) \leq \frac{\sigma^2}{c}$, for $\forall t \geq 0$ (Alistarh et al., 2018; Raghu Meka, 2019).

Since, $\|\nabla E(W_t)\|^2 = \|\mathbb{E}(\bar{v}_t)\|^2 = \mathbb{E}(\|\bar{v}_t\|^2) - Var(\bar{v}_t)$

$$\mathbb{E}(E(W_{t+1})) \leq \mathbb{E}(E(W_+)) + \mathbb{E}(\langle \mathbb{E}(\bar{v}_t), W_t - W_+ \rangle) - \frac{\eta c}{2}(\mathbb{E}(\|\bar{v}_t\|^2) - Var(\bar{v}_t)) + \frac{\eta}{2}\sigma^2$$

$$= \mathbb{E}(E(W_+)) + \langle \mathbb{E}(\bar{v}_t), \mathbb{E}(W_t - W_+) \rangle - \frac{\eta c}{2}\mathbb{E}(\|\bar{v}_t\|^2) + \eta\sigma^2$$

Using Cauchy Swartz inequality as described in (Bubeck, 2015):

$$\mathbb{E}(E(W_{t+1})) \leq E(W_+) + \|\bar{v}_t\| \|W_t - W_+\| - \frac{\eta c}{2}\|\bar{v}_t\|^2 + \eta\sigma^2$$

$$= E(W_+) + \frac{1}{2\eta c}(-(\|W_t - W_+\| - \|\eta c v_t\|)^2 + \|W_t - W_+\|^2) + \eta\sigma^2$$

$$\leq E(W_+) + \frac{1}{2\eta c}(\|W_t - W_+\|^2 - (\|W_{t+1} - W_+\|)^2) + \eta\sigma^2$$

For $t = 0$, $E(W_1) \leq E(W_+) + \frac{1}{2\eta c}(\|W_0 - W_+\|^2 - \|W_1 - W_+\|^2) + \eta\sigma^2$

For $t = T - 1$ (where $T > N$), $E(W_T) \leq E(W_+) + \frac{1}{2\eta c}(\|W_{T-1} - W_+\|^2 - \|W_T - W_+\|^2) + \eta\sigma^2$

Sum up $t = 0 \ldots (T-1)$:

$$\sum_{t=1}^{T} E(W_t) \leq TE(W_+) + \frac{1}{2\eta c}(\|W_0 - W_+\|^2 - \|W_T - W_+\|^2) + T\eta\sigma^2$$

$$\leq TE(W_0) + \frac{1}{2\eta c}\|W_0 - W_+\|^2 + T\eta\sigma^2$$

$$\frac{1}{T}\sum_{t=1}^{T} E(W_t) \leq E(W_+) + \frac{1}{2\eta cT}\|W_0 - W_+\|^2 + \eta\sigma^2 \qquad \blacksquare$$

Rearranging the variables would finally provide the convergence term as shown in Eq. (3):

$$\mathbb{E}\left(E(W) - \min_{W} E(W)\right) \leq \frac{1}{2\eta cT}\left\|\operatorname*{argmax}_{W} E(W) - \operatorname*{argmin}_{W} E(W)\right\|^2 + \eta\sigma^2 \qquad (3)$$

So the convergence rate is of the order $O\left(\frac{1}{cT} + \sigma^2\right)$. This shows that both SGD and parallel SGD eventually converge to $O\left(\frac{1}{T} + \sigma^2\right)$, however, the number of CPUs matters in the early stage of training and speed up the convergence.

In the asynchronous case, faster threads do not wait for slower threads to finish and continue to calculate new gradients. An assumption is that a slave thread receives updated weight on request rather than instantly after the update.

If the next iteration is considered as the completion of the slowest thread, then within one iteration there are $M$ smaller steps $t + s1, t + s2, \ldots, t + sM - 1$. The combinations of steps for one iteration can be written in a generalized form as:

$$W_{t+1} = W_t - \eta \sum_{j=1}^{c_0} v_t(W_t, d_j) - \sum_{j=1}^{c_1} v_t(W_{t+S1}, d_{c_1+j}) - \cdots \sum_{j=1}^{c_M} v_t(W_{t+sM}, d_{c_M+j})$$

where $\sum_{i=0}^{M} c_i \geq M$ and $c_i \leq M$

For simplicity, $W_{t+1} = W_t - \eta c_0 \bar{v}_t - \eta c_1 \bar{v}_{t+s1} - \cdots - \eta c_M M \bar{v}_{t+sM}$

Using expected value both sides:

$$\mathbb{E}(W_{t+1}) = \mathbb{E}(W_t) - \eta c_0 \mathbb{E}(\bar{v}_t) - \cdots - \eta c_M \mathbb{E}(\bar{v}_{t+sM})$$

Since expected value of a sample mean is population mean, therefore

$$\mathbb{E}(W_{t+1}) = \mathbb{E}(W_t) - \eta c_0 \nabla E(W_t) - \cdots - \eta c_M \nabla E(W_{t+sM})$$

Or, $\mathbb{E}(W_{t+1}) = \mathbb{E}(W_t) - \eta c_0 \nabla E(W_t) - \cdots - \eta c_M \nabla E(W_{t+sM})$

$$= \mathbb{E}(W_t) - \eta \sum_{i=0}^{M} c_i \nabla E(W_{t+si}) = \mathbb{E}(W_t) - \eta M \sum_{i=1}^{M} \frac{c_i}{M} \nabla E(W_{t+si})$$

$\frac{c_i}{M}$ can be rightly treated as a probability for the occurrence of gradient $\nabla E(W_{t+si})$.

Or simply: $\mathbb{E}(W_{t+1}) = \mathbb{E}(W_t) - \eta M \mathbb{E}(\nabla E(W_t))$

$W_{t+1} = W_t - \eta M \nabla E(W_t)$

The rest of the proof is the same as SSGD. Finally the convergence term can be given as:

$$\mathbb{E}\left(E(W_t) - \min_W E(W)\right) \leq \frac{1}{2\eta MT} \left\| \operatorname*{argmax}_W E(W) - \operatorname*{argmin}_W E(W) \right\|^2 \qquad (4)$$

So the convergence rate is of the order $O\left(\frac{1}{MT}\right)$ which is better than SSGD of $O\left(\frac{1}{cT} + \sigma^2\right)$ for real time systems, otherwise asymptotically both are the same.

## 4. Proposed Guided approach with gSGD for parallel GD

Guided SGD (gSGD) is a recently proposed gradient descent algorithm in (Sharma, 2018) that identifies consistent data samples while training, to boost the classification accuracy. It was reported that the classification accuracy was improved by approximately 3% for some datasets in a limited time budget. gSGD is also compatible with all the other popular variations of SGD such as RMSprop (Zeiler, 2012) and Adagrad (Duchi et al., 2011).

Inconsistent data instances are simply the data instances within the neighborhood of instance $j$; which individually performs better, while the average error value $\bar{E}_t$ performs worse than the average error of the previous iteration $\bar{E}_{t-1}$, and vice-versa. The brief pseudocode of the algorithm is given in Fig. 5 where the algorithm begins by initializing weight vector ($W_0$), learning rate ($\eta$), total iterations ($T$) and neighborhood size ($\rho$) (Sharma, 2018). Neighborhood allocation allows gSGD to run through $\rho$ training examples recursively before further refinement of $W$ is done with the consistent data. Consistent data is selected from $\rho$ training examples at a time. Each data instance $d_i$ is randomly selected to update the weight vector ($W_t$) at iteration $t$. The neighborhood of an instance $i$ is kept in sets $\psi$ where they track consistent instances. After collecting the data about inconsistency for $\rho$ iterations, only consistent instances are extracted and kept in $\psi$. Lastly, $W_t$ is further updated with consistent instances.

```
//Initialize η, ρ, T and W₀
For t = 1…T
    Wₜ = Wₜ₋₁ − ηv(dᵢ)
    Ēₜ = calcAvgError();
    ψ = collectConsistentBatches(d₁,…,d_N);
    If t mod ρ = 0, i.e., max delay tolerance (ρ) is reached
        ψ = getMostConsistentBatches(ψ, Ēₜ);
        For j = 1:‖ψ‖
            Wₜ = Wₜ − ηv(ψⱼ)
        End for
    End If
End For
```

**Fig. 5.** Pseudocode for the original gSGD

Its flowchart is given in Fig. 6 where the algorithm starts with a random selection of a data instance whose gradient is computed to update the weight vector. After $\rho$ iterations, weight vector is further refined with consistent neighboring data instances. $\rho$ is also the neighborhood size which is generally assigned to a constant value 10.

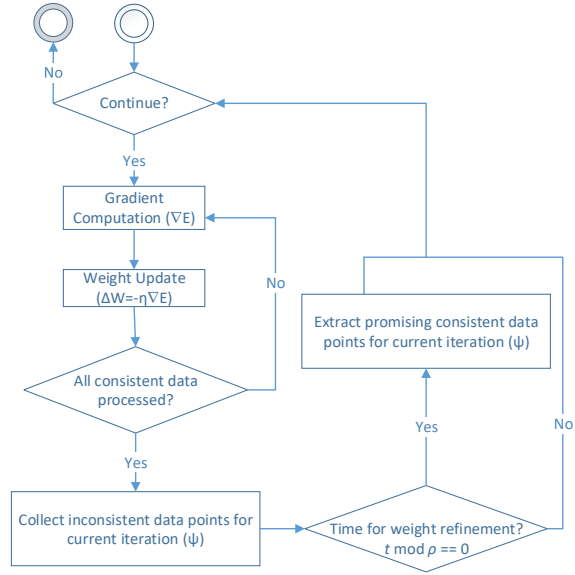

**Fig. 6.** Flowchart of gSGD

The guided approach for SGD can be similarly incorporated in both SSGD and ASGD which can be represented as gSSGD and gASGD respectively or gS/ASGD collectively. We are using a simplified variation of gSGD that is more efficient and works well on mini-batch GD has been discussed next.

The first step towards the compensation of delay is to define a threshold ($\rho$) for delay tolerance in the weight update, otherwise, the gradient descent becomes unpredictable. The guided approach works on this threshold to overcome the deviation, by identifying individual delays as consistent or inconsistent. As discussed, consistency for an individual data example is based on its direction of gradient for a loss function that corresponds to the true gradient and results in a gradient descent (convergence), and vice versa for inconsistency. The consistent data examples help in steering the convergence towards gradient descent.

The algorithm for parallel guided GD, gS/ASGD given in Fig. 7 adds parallelization to the original gSGD given in Fig. 5. Its gradient computation section is not different from S/ASGD, however, the parameter server runs the consistent data after every $\rho$ updates. Computing training loss is an expensive task that's why a small validation set is randomly picked (called verification data) to estimate the training loss $\bar{E}_t$. $\psi$ represents individual error loss of previous two mini-batches in every iteration which is then sorted after $\rho$ iterations to get the most consistent mini-batches. The size of the most consistent mini-batches is generally not more than 4 to keep the algorithm efficient. Lastly, weight $W$ is updated with the most consistent mini-batches. The code for this algorithm is available at [https://github.com/anuraganands – after acceptance].

```
//Initialize η, ρ, T and W₀
Listen
   If a gradient v(dᵢ) received
      Wₜ = Wₜ₋₁ − ηv(dᵢ)
      Ēₜ = approximateAvgError();
      ψ = collectConsistentBatches(dᵢ, dᵢ₋₁, dᵢ₋₂);
      If t mod ρ = 0, i.e., max delay tolerance (ρ) is reached
         ψ = getMostConsistentBatches(ψ, Ēₜ);
         For i = 1:‖ψ‖
            Wₜ = Wₜ − ηv(ψᵢ)
         End for
      End If
   End If
While (t = 1: T)
```

**Fig. 7.** Algorithm for the parameter server using the guided approach

The analysis of the convergence rate for gS/ASGD is similar to S/ASGD. The weight update in gSSGD is: $W_{t+1} = W_t - \eta(c + \frac{c'}{\rho})\bar{v}_t$ which would result in a similar convergence rate where $c' < c$ is the size of additional consistent data samples runs after every $\rho$ iterations. So its convergence rate would be same as SSGD: $O\left(\frac{1}{cT} + \sigma^2\right)$ as $c + \frac{c'}{\rho} \leq 2c$. We used $c = \rho$ in our experiments so its convergence rate can also be given as $O\left(\frac{1}{\rho T} + \sigma^2\right)$. Similarly, the convergence rate of gASGD would be same as ASGD. Note that the worst time complexity for the gSGD is also same as SGD according to (Sharma, 2018).

5. **Experiments & Discussion**

A comprehensive analysis was performed to analyze the impact of the guided approach on S/ASGD. The coding was completely written in Java and Matlab. Java is used to spawn the threads and Matlab is used to execute the algorithms. To establish the proof of concept we have conducted the experiment on logistic regression only in this paper. Theoretically, our approach can be easily extended to other GD based learning systems such as recurrent neural network and convolutional neural network for bigger datasets as only GD needs to be monitored. Similarly, other variations of GD such as RMSprop (Zeiler, 2012) and Adam (Kingma and Ba, 2014) can also be used as demonstrated in (Sharma, 2018).

## 5.1 Test Data and Parameter Settings

All experiments were conducted on 24 Intel® Xeon® E5-4610 CPUs in a single server. We used the recommended parameter settings from (Sharma, 2018) shown in Table 1.

Table 1. Parameter setting for gSGD and SGD

| Parameter | Value |
|---|---|
| Epochs | 50 |
| Attempts | 30 consecutive runs |
| Training:Testing | 80:20 (of entire data examples) |
| Training:Validation | 80:20 (of training data ) |
| Learning rate ($\eta$) | 0.2 |
| Delay tolerance ($\rho$) | 10 |

The major focus of the experiment is to demonstrate the applicability of our proposed guided approach in delay compensation due to parallel gradient computations. We have employed 9 benchmark datasets from the UCI Library (Dheeru and Karra Taniskidou, 2019) to have comparative analysis of parallel SGD algorithms on classification accuracy. All these datasets have either two or three classes. No data cleaning or preprocessing techniques have been applied to the datasets unless otherwise specified. *Pima Indian diabetes* and *liver disorder* datasets are found to be very noisy, therefore noise filtering through outlier detection and removal has been applied. Statistical inter-quartile range (IQR) outlier detection method available in (Eibe Frank et al., 2016) has been used to remove stochastic noise from these two datasets. This technique has been commonly used in the medical domain with increased descriptive classification accuracy but with low predictive accuracy (Laurikkala et al., 2000; Solberg and Lahti, 2005).

## 5.2 Experimental Results

Table 2 and Table 3 show the best and average classification accuracies attained for each dataset after 30 runs respectively. Datasets with 'filtered' suffix indicates IQR outlier detection and removal have been applied. The experiment is grouped into three sets: SGD, SSGD, and ASGD. Each set is tested with and without the guided approach. The guided approach is prefixed with small 'g' with the algorithms' name. The bold font in the table indicates the better approach within the set and the highlighted cells show the overall best. The average result is reported with tolerance values in Table 3. Tolerance value is computed by collecting the classification accuracy of 30 runs in a sorted order. Thereafter the outliers are removed by collecting results from the first to the third quartile only. Tolerance is half of the difference between these two quartiles.

**Table 2.** Best Classification Accuracies – Canonical GD

| Datasets | Algorithms | | | | | |
|---|---|---|---|---|---|---|
| | SGD | gSGD | SSGD | gSSGD | ASGD | gASGD |
| Pima Indian diabetes | 80.13 | **81.11** | 77.20 | **81.43** | 78.18 | **81.11** |
| Pima Indian diabetes (filtered) | 79.44 | **82.23** | 75.61 | **79.43** | 74.56 | **77.00** |
| Breast Cancer Diagnostic | 98.24 | **98.68** | 95.60 | **97.80** | **96.92**† | 95.60† |
| Haberman | 81.97† | 81.97† | 81.97† | **81.97**† | 81.97† | 81.97† |
| Liver Disorder | 71.01 | **73.91** | 69.57 | **73.19** | 68.84 | **69.57** |
| Liver Disorder (filtered) | 76.74 | 76.74 | 69.77 | **74.42** | **68.99** | 68.22 |
| New-thyroid | 98.84 | 98.84 | 82.56 | **89.54** | 82.56 | 82.56 |
| Cancer | 99.45 | 99.45 | 98.90 | 98.90 | 98.90† | 98.90† |
| Phishing | 84.29 | **84.84** | 87.25 | **87.62** | 86.32 | **87.62** |

**Table 3.** Average Classification Accuracies – Canonical GD

| Datasets | Algorithms | | | | | |
|---|---|---|---|---|---|---|
| | SGD | gSGD | SSGD | gSSGD | ASGD | gASGD |
| Pima Indian diabetes | 76.1±1.5 | **77.7±1.5** | 71.7±2.0 | **73.8±1.5** | **71.8±2.1** | 71.3±2.0 |
| Pima Indian diabetes (filtered) | 74.9±1.4 | **76.8±1.2** | 71.6±1.9 | **74.6±1.4** | 71.1±1.4 | **72.3±0.9** |
| Breast Cancer Diagnostic | 95.8±1.1 | **96.9±0.9** | 93.6±0.7 | **95.2±0.9** | 93.6±1.1† | 93.6±1.1† |
| Haberman | 74.6±2.5 | **77.5±2.0**† | 73.8±2.5 | **73.8±2.5**† | 74.2±2.0† | 74.2±2.0† |
| Liver Disorder | 64.9±2.5 | **66.3±2.5** | 62.0±3.3 | **62.7±2.5** | **62.0±2.5** | 60.9±2.9 |
| Liver Disorder (filtered) | 69.4±2.7 | **71.3±2.3** | 60.9±2.7 | **67.8±1.9** | **62.4±2.7** | 61.6±4.3 |
| New-thyroid | 92.4±2.9 | **96.5±2.3** | 73.3±3.5 | **79.1±3.5** | 73.3±4.7 | 73.3±3.5 |
| Cancer | 97.8±0.5 | 97.8±0.5 | 97.0±0.8 | **97.5±0.8** | 97.0±0.8† | 97.0±0.8† |
| Phishing | 82.2±1.0 | **82.3±1.3** | 82.7±0.8 | **83.8±0.8** | 83.2±0.7 | 83.2±0.9 |

Non-parametric test was also performed with Wilcoxon test for ranking (two-tailed) to show null hypothesis with $p > 0.05$ (there is no significance difference between tested algorithms) is wrong. The test is conducted on all three groups of algorithms: SGD & gSGD, SSGD & gSSGD, and ASGD & gASGD. Insignificant difference is denoted by † in Table 2 and Table 3. Most of the test cases except *Haberman* dataset show there is a significant difference for synchronous and naïve approaches with their guided counterpart. That means guided approach has been able to make a significant improvement. On the other hand, guided approach has not been able to significantly impact the asynchronous approach. The results of *Haberman*, *Breast Cancer Diagnostic* and *Cancer* datasets are not even significantly different, possibly due to less inconsistency and not capitalizing the filtering process due to the asynchronous selection and irregular updates with gradient values adding up the variance. Fig. 8 also demonstrates that the above mentioned datasets *Breast Cancer Diagnostic* and *Cancer* are much smoother (have less inconsistency) compared to others. *Haberman*, however, is too noisy to be improved. All these three datasets where gASGD and ASGD have shown the results as statistically insignificant, are shown with † symbol in the legend of the figure.

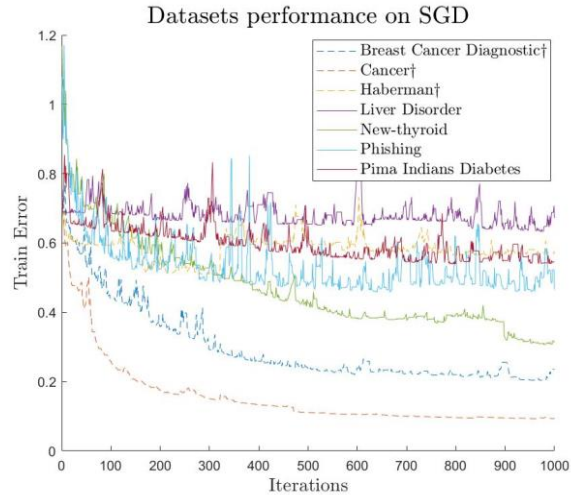

**Fig. 8.** Training of SGD on various datasets

The experimental results show that the guided approach mostly improves the solution. Naïve synchronous and asynchronous approaches have produced similar results, however, the synchronous approach has been marginally better in 7/9 datasets. The overall performance comparison of parallel SGD algorithms on classification accuracy is shown in Fig. 9. The dataset names have been given in short-form in the horizontal axis. '*' means the filtered dataset. gSSGD has turned out to be the most promising parallel algorithm while others have produced similar results. The guided approach works convincingly well with SSGD. It makes a significant improvement of up to 7% (for *New-thyroid*) with its delay compensation strategy of the guided approach.

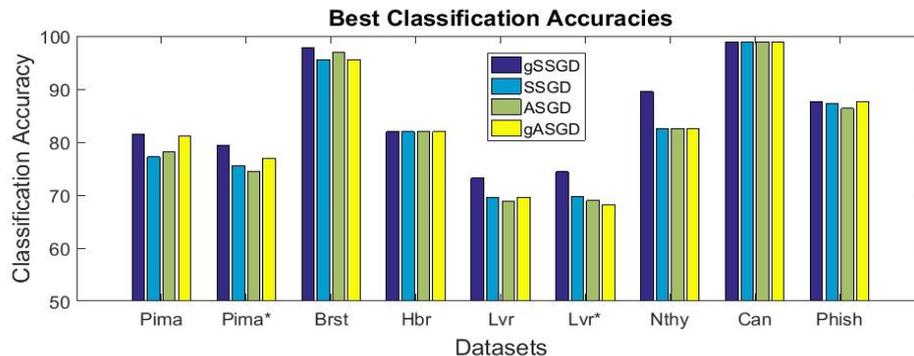

**Fig. 9.** Overall comparison of performance on classification accuracy for parallel SGD algorithm

The sequential gSGD has been the most prominent algorithm of the entire test set while gSSGD has been the best among parallel SGD algorithms. With this delay compensation, there is an attempt to bring the classification accuracy close to the sequential SGD/gSGD. The performance comparison graphs in Fig. 10 illustrate how gSSGD has fared with other algorithms. The vertical axis represents the difference in the classification accuracy of gSSGD with other algorithms. The positive values show gSSGD is better and vice versa for negative value. gSSGD clearly outperforms other parallel algorithms namely SSGD and ASGD. The improvement is as much as 7%. It also catches up reasonably well with the results of sequential SGD and gSGD where it dominates in 3/9 and 2/9 datasets respectively, however, it is expected for sequential SGD/gSGD to perform better.

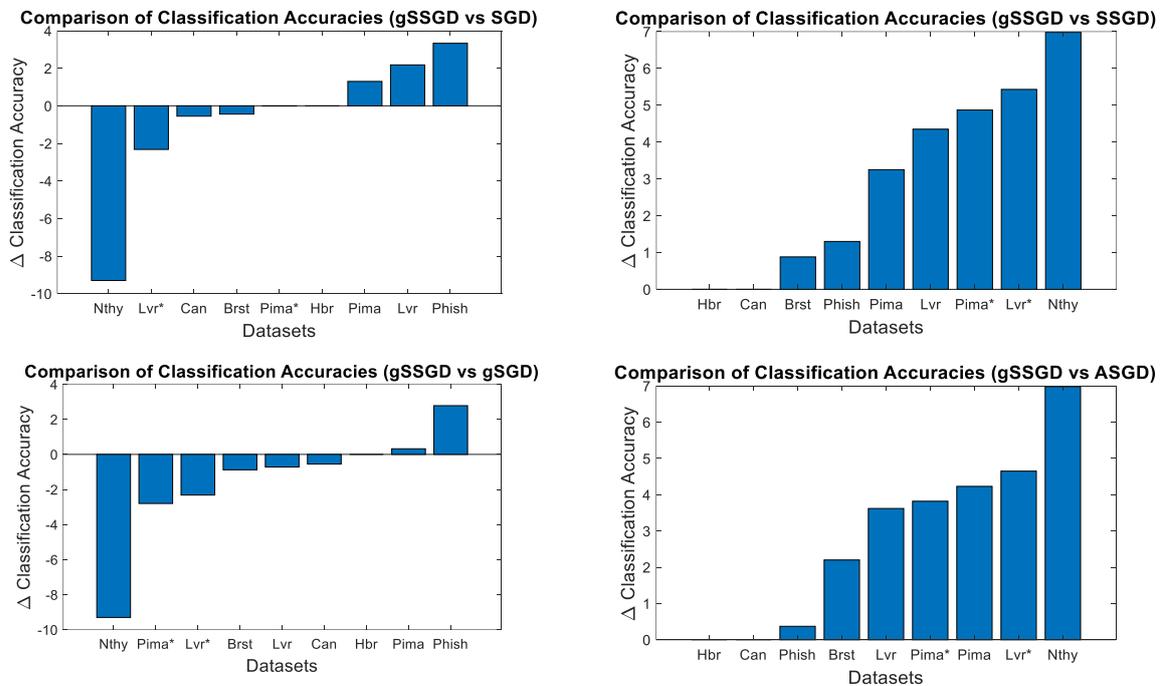

**Fig. 10.** Comparison of classification accuracies for gSSGD

It was demonstrated in (Sharma, 2018) that the guided approach works on most of the known variations of SGD. Its versatility and the outstanding performance of synchronous SGDs solicits further tests in parallel environment for some other popular variations of SGD such as RMSprop (Zeiler, 2012) and Adagrad (Duchi et al., 2011). Since the guided approach uses the same gradient computation formula, it has been easily incorporated into RMSprop and Adagrad. Only the canonical weight update section of the original algorithm in Fig. 7 needs to be replaced with the chosen variation of weight update. For example, the weight update for RMSprop is $W_t =$ $W_{t-1} - \eta \frac{v}{\sqrt{(r_t+\varepsilon)}}$, where $\varepsilon = 1.0 \times 10^{-8}$, and $r_t = \beta r_{t-1} + (1-\beta)v_t^2$ where $r_1 = v_1$ and $\beta = 0.9$. The complete algorithm for the parallelized guided version of RMSprop's parameter server is given in Fig. 11. Similarly, Adagrad has been coded for the parallel environment.

```
//Initialize η, ρ, T and W₀
//Initialize β = 0.9 //RSMprop specific constant
Listen
    If a gradient v(dᵢ) received
      rₜ = βrₜ₋₁ + (1 − β)v(dᵢ)²
      Wₜ = Wₜ₋₁ − ηv(dᵢ)/√(rₜ+ε)  //RMSprop weight update
      Ēₜ = approximateAvgError();
      ψ = collectConsistentBatches(dᵢ,dᵢ₋₁,dᵢ₋₂);
      If t mod ρ = 0, i.e., max delay tolerance (ρ) is reached
        ψ = getMostConsistentBatches(ψ,Ēₜ);
        For i = 1:‖ψ‖
          Wₜ = Wₜ − ηv(ψᵢ)/√(rₜ+ε) //RMSprop weight update
        End for
      End If
    End If
While (t = 1: T)
```

**Fig. 11.** Algorithm for the parameter server for RMSprop

The synchronous RMSprop and Adagrad are now called SRMSprop and SAdagrad respectively. Using the same parameter settings given in Table 1, both of these algorithms have been tested on all 9 datasets. The experimental results in Table 4 and Table 5 show the best and average classification accuracies attained for each dataset after 30 runs respectively. The same Wilcoxon significance testing has been conducted on guided versus original algorithm where the insignificant difference is denoted by † (if any). Considering the best classification accuracies, gSRMSprop outperforms SRMSprop in 6/9 datasets whereas gSRMSprop has completely dominates SRMSprop for 8/9 datasets for the average classification accuracy. Likewise, gSAdagrad has completely outperformed SAdagrad on average classification but showed only minor improvement on 3/9 datasets. This demonstrates that guided approach certainly enhances the performance of SSGD and its variations on parallel environment as well. It can also be observed that gSRMSprop even further improves the classification accuracy as shown by the highlighted cells in Table 4 and Table 5. Some of its noteworthy results are for *cancer* and *breast cancer diagnostic* where the classification accuracy has clocked almost 100%.

Table 4. Best Classification Accuracies – SSGD, SRMSprop and SAdagrad

| Datasets | Algorithms | | | | | |
|---|---|---|---|---|---|---|
| | SSGD | gSSGD | SRMSprop | gSRMSprop | SAdagrad | gSAdagrad |
| Pima Indian diabetes | 77.20 | **81.43** | 80.78 | **81.43** | 71.01 | **73.94** |
| Pima Indian diabetes (filtered) | 75.61 | **79.43** | 79.44 | **80.84** | 72.47 | **74.22** |
| Breast Cancer Diagnostic | 95.60 | **97.80** | 97.80 | **99.12** | 95.60 | 95.60 |
| Haberman | 81.97† | **81.97**† | **82.79** | 81.97 | 81.97 | 81.97 |
| Liver Disorder | 69.57 | **73.19** | 75.36 | **78.26** | **70.29** | 69.57 |
| Liver Disorder (filtered) | 69.77 | **74.42** | **77.52** | 76.74 | 75.19 | 75.19 |
| New-thyroid | 82.56 | **89.54** | **100.0** | **100.0** | 74.42 | **89.54** |
| Cancer | 98.90 | 98.90 | 98.90 | **99.45** | 98.90 | 98.90 |
| Phishing | 87.25 | **87.62** | 87.25 | **87.43** | 86.14 | 86.14 |

Table 5. Average Classification Accuracies – SSGD, SRMSprop and SAdagrad

| Datasets | Algorithms | | | | | |
|---|---|---|---|---|---|---|
| | SSGD | gSSGD | SRMSprop | gSRMSprop | SAdagrad | gSAdagrad |
| Pima Indian diabetes | 71.7±2.0 | **73.8±1.5** | 75.1±2.1 | **77.5±1.3** | 67.3±1.8 | **68.2±2.1** |
| Pima Indian diabetes (filtered) | 71.6±1.9 | **74.6±1.4** | 75.4±1.6 | **78.2±0.9** | 66.6±2.1 | **68.1±2.3** |
| Breast Cancer Diagnostic | 93.6±0.7 | **95.2±0.9** | 96.3±0.7 | **98.0±0.7** | 91.9±1.1 | **92.7±1.1** |
| Haberman | 73.8±2.5 | **73.8±2.5**† | 75.8±2.0 | **77.0±1.6** | 74.2±2.0 | **74.6±2.5** |
| Liver Disorder | 62.0±3.3 | **62.7±2.5** | 67.8±1.8 | **70.7±1.8** | 60.9±3.6 | **61.6±3.6** |
| Liver Disorder (filtered) | 60.9±2.7 | **67.8±1.9** | 70.9±3.5 | **72.1±2.3** | 61.2±3.9 | **61.6±4.3** |
| New-thyroid | 73.3±3.5 | **79.1±3.5** | 95.9±1.7 | **97.7±1.2** | 78.5±1.7 | **80.2±3.5** |
| Cancer | 97.0±0.8 | **97.5±0.8** | 98.1±0.8 | 98.1±0.8 | 97.3±0.5 | 97.3±0.5 |
| Phishing | 82.7±0.8 | **83.8±0.8** | 83.7±0.7 | **84.0±0.8** | 82.2±0.8 | **82.9±1.0** |

## 5.3 Parameter Analysis

As discussed earlier, gSGD approach of tracking inconsistent data from (Sharma, 2018) has been used as a delay compensation strategy for the parallel algorithms. The delay tolerance threshold ($\rho$) determines the number of parallel processes. Technically, $\rho$ can have any value $\geq 1$ which can enhance the processing speed by $\sim\rho$ fold if inter-process communication overhead is ignored. The impact of different $\rho$ values on the classification accuracy tested on a selected dataset New-thyroid dataset is shown in Fig. 12. The $\rho$ value with the percentage of training data size is also given. As the $\rho$ value increases the accuracy diminishes. $\rho = 0$ shows a sequential process with no delay to compensate. The experiment on another dataset *Breast Cancer Diagnostic* also shows the similar results in Fig. 13.

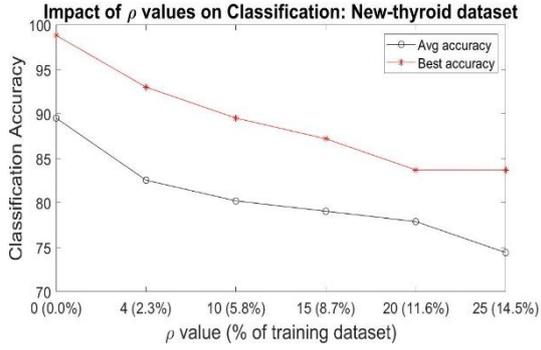

**Fig. 12.** Impact of $\rho$ on classification accuracy for New-thyroid dataset

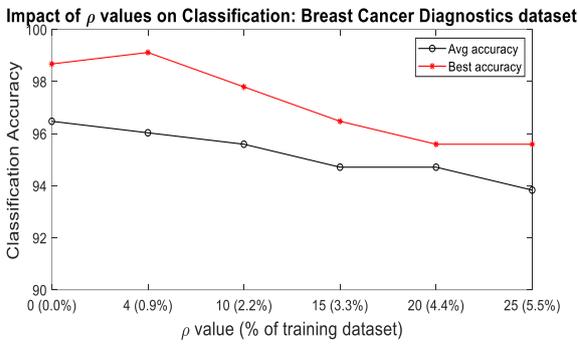

**Fig. 13.** Impact of $\rho$ on classification accuracy for Breast Cancer Diagnostic dataset

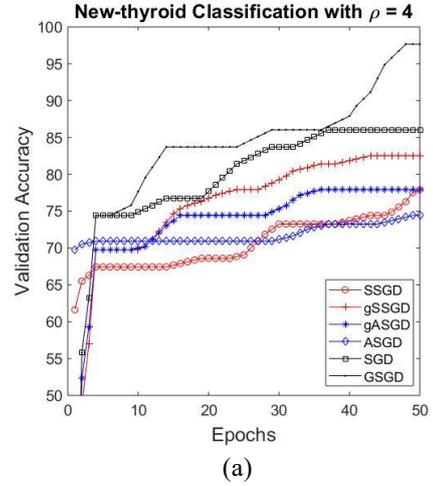

(a)

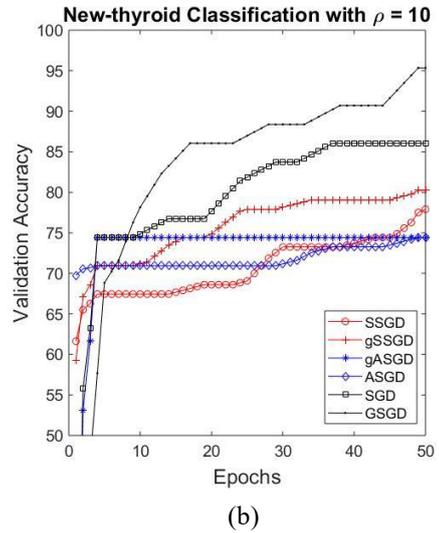

(b)

**Fig. 14.** Validation accuracy progression for all parallel and sequential SGD algorithms on New-thyroid dataset

The progression of validation accuracy for all the tested sequential and parallel algorithms for the dataset New-thyroid has been given in Fig. 14. Indeed, the sequential algorithms especially gSGD clearly outweigh parallel algorithms in classification accuracy. Nevertheless, the delay compensation with $\rho = 4$ and 10 shows significant improvement of classification accuracy in SSGD. Appropriate delay compensation is necessary to produce a quality solution in quick time using a parallel approach. However, some compromise has to be made in terms of classification accuracy. So the $\rho$ value needs to be chosen wisely considering its convergence rate of $O\left(\frac{1}{\rho T} \pm \sigma^2\right)$. The higher the $\rho$ the faster the convergence but the poorer the classification accuracy.

## 6. Conclusion

This paper shows the proof of the concept of parallelization of SGD with a guided approach to logistic regression. The guided approach has been used to compensate for the delay caused by parameter updates in parallel processing. The synchronous gSSGD mitigates the impact of delay with a reasonable efficiency and same convergence rate of naïve SSGD $O\left(\frac{1}{cT} \pm \sigma^2\right)$. gSSGD also outperforms naïve SSGD in classification accuracy in 8/9 datasets where improvement of up to 7% has been observed. Notably, the classification accuracy produced by gSSGD is comparative to the sequential SGD as well where it has even produced better results in 3/9 datasets. The immediate future work is to apply gSSGD on deep networks such as Convolutional Neural Networks. It can also be applied to other known delay compensation approaches on parallel SGD such as DC-ASGD (Zheng et al., 2017).